%% file: nips_2018.tex
\documentclass{article}

\usepackage[final]{nips_2018}

\usepackage[utf8]{inputenc} 
\usepackage[T1]{fontenc}    
\usepackage{hyperref}       
\usepackage{url}            
\usepackage{booktabs}       
\usepackage{amsfonts}       
\usepackage{nicefrac}       
\usepackage{microtype}      
\usepackage{amsmath,amssymb}
\usepackage{xcolor}
\usepackage{bbm}
\usepackage{multirow}
\usepackage{graphicx}
\usepackage[export]{adjustbox}[2011/08/13]
\usepackage{caption}
\usepackage{subfig}
\usepackage{float}
\usepackage{colortbl}

\newcommand{\setmode}[1]{\def\mode{#1}}
\setmode{final} 
\long\def\IGNORE#1{} \long\def\COMMENT#1{}
\def\authornote#1#2#3{{\color{#2}{\textsl{\small#1:[*#3*]}}}}

\ifthenelse{\equal{\mode}{draft}} { 
    \newcommand{\yhnote}[1]{\authornote{YH}{red}{#1}} 
    \newcommand{\zk}[1]{\authornote{ZK}{blue}{#1}} 

    } {}

\ifthenelse{\equal{\mode}{final}} {
    \newcommand{\advise}[1]{}
    \newcommand{\yhnote}[1]{}
    \newcommand{\zk}[1]{}
    \typeout{************* MODE: Final}
    } {}

\makeatletter
\newcommand*{\centerfloat}{%
  \parindent \z@
  \leftskip \z@ \@plus 1fil \@minus \textwidth
  \rightskip\leftskip
  \parfillskip \z@skip}
\makeatother

\title{Re-evaluating Continual Learning Scenarios: A Categorization and Case for Strong Baselines}

\newcommand*{\affaddr}[1]{#1} 

\newcommand*{\email}[1]{\texttt{#1}}

\author{%
Yen-Chang Hsu, Yen-Cheng Liu, Anita Ramasamy, and Zsolt Kira\\
\affaddr{Georgia Institute of Technology}\\
\email{\{yenchang.hsu,ycliu,anita.ramasamy,zkira\}@gatech.edu}\\%
}

\begin{document}
\setcitestyle{numbers}

\maketitle

\begin{abstract}

Continual learning has received a great deal of attention recently with several approaches being proposed. However, evaluations involve a diverse set of scenarios making meaningful comparison difficult. This work provides a systematic categorization of the scenarios and evaluates them within a consistent framework including strong baselines and state-of-the-art methods. The results provide an understanding of the relative difficulty of the scenarios and that simple baselines (Adagrad, L2 regularization, and naive rehearsal strategies) can surprisingly achieve similar performance to current mainstream methods. We conclude with several suggestions for creating harder evaluation scenarios\yhnote{more complex multi-dataset evaluation : We don't have the analysis in the main text. Is it appropriate to mention it in abstract?} and future research directions. The code is available at \href{https://github.com/GT-RIPL/Continual-Learning-Benchmark}{https://github.com/GT-RIPL/Continual-Learning-Benchmark}.

\end{abstract}

\section{Introduction}

While current learning-based methods can achieve high performance on tasks, they only perform well when the testing data is similarly distributed as the training data. In other words, they cannot adapt continuously in dynamic environments where situations can significantly change. Such adaptation is desirable for any intelligent system, and is the hallmark of learning in biological systems. One approach to this problem is continual learning, where models are updated incrementally as data streams in. However, deep neural networks, which are currently state of art for many applications, are known to suffer catastrophic interference or forgetting when incrementally updated through gradient-based methods. This leads to the model forgetting how to solve old tasks after being exposed to a new one due to interference caused by  parameter updates. 

To address this problem, several approaches have been proposed and a number of experimental methodologies (i.e. datasets, learning curricula, and architectures) have been used for evaluation. In this paper, we argue that the current set of evaluations have significant limitations, including lack of uniformity across the different experimental methodologies, lack of hyper-parameter tuning of reasonable baselines under similar tuning budgets as the proposed methods, and simplicity of the tasks (e.g. short task queues). Towards this end, we make several contributions in this paper: 1) A categorization of a large number of experimental methodologies into a few canonical settings along with a comparison of their difficulty, 2) A uniform but flexible framework for generating scenarios under this categorization and systematic evaluation of current state of art methods, and 3) Demonstration that very simple baselines can be surprisingly effective if used properly and result in comparable or better performance against the current state of art. We have released our framework (written in PyTorch) to enable fair and uniform evaluation to aid the community, and conclude with some suggested modifications to the scenarios to increase realism of the evaluation.


\begin{table}[]
\centering
\caption{The continual learning scenarios categorized by the difference between the old and new task. $P(X)$: The distribution of input data. $P(Y)$: The distribution of target labels. $\{Y_1\} \neq \{Y_2\}$: The labels are from a disjoint space which is differentiated by task identity. S: Single-headed model. M: Multi-headed model. I: Known task identity.}
\label{tbl:task_dif}
\begin{tabular}{c|ccc|l}
\toprule
\multirow{2}{*}{Learning scenario} & \multicolumn{3}{c|}{Old Task ($T_1$) versus New Task ($T_2$)} & \multirow{2}{*}{Remark} \\
                                   & $P(X_1) \neq P(X_2)$    & $P(Y_1) \neq P(Y_2)$    & $\{Y_1\} \neq \{Y_2\}$    &                        \\ \midrule
Non-incremental learning           &                  &                &                  &                                            \\
Incremental domain learning        & \checkmark       &                &                  & S                         \\
Incremental class learning         & \checkmark       & \checkmark     &                  & S                         \\
Incremental task learning          & \checkmark       & \checkmark     & \checkmark       & M, I    \\
\bottomrule
\end{tabular}
\end{table}

\begin{figure}[]
  \centering
  \includegraphics[clip, trim=0cm 1.3cm 0cm 1.7cm, width=1\textwidth]{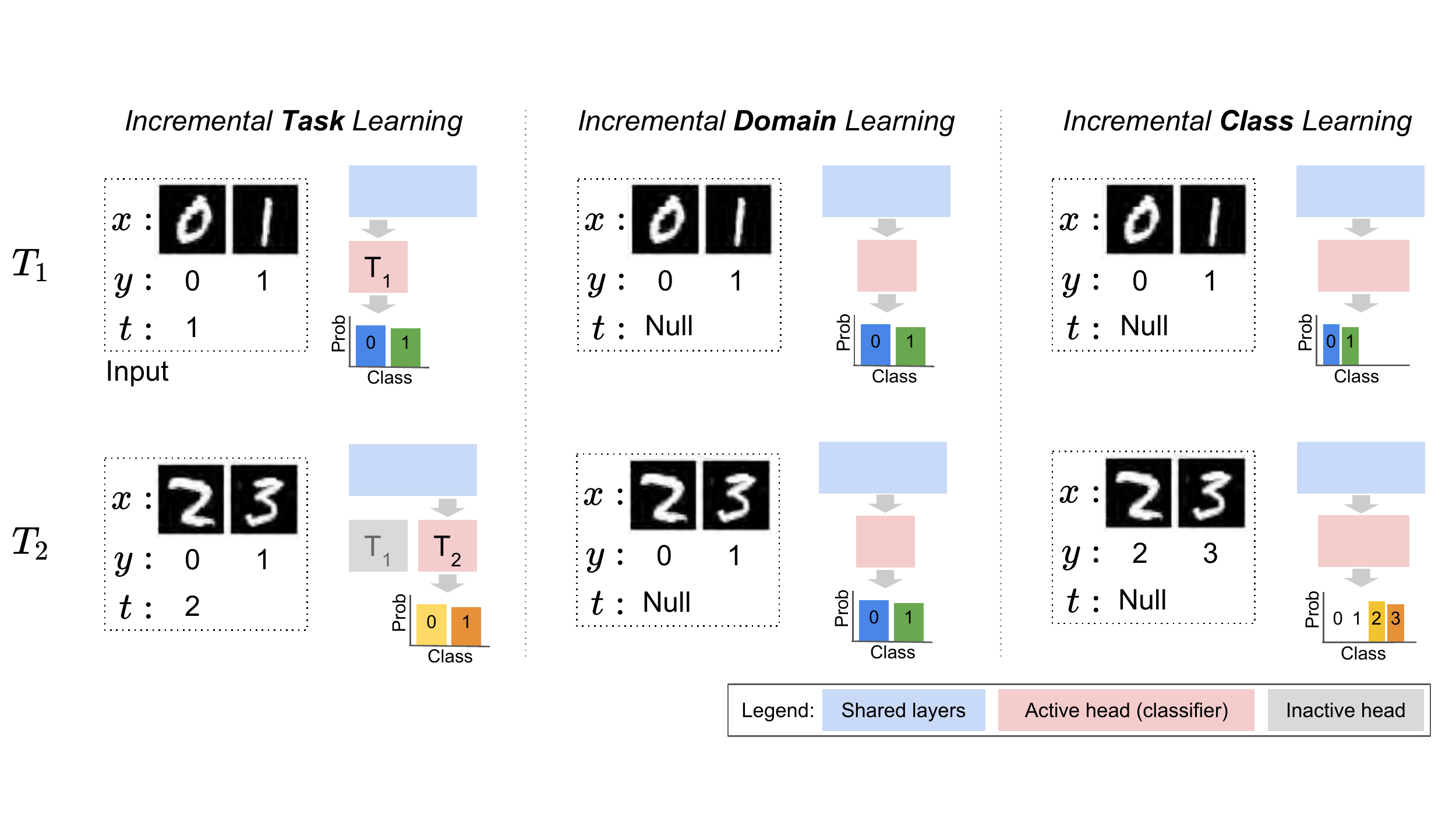} 
  \caption{The three continual learning scenarios generated by Split MNIST. In each sub-figure, the left dotted rectangle represents the inputs for training, in that $(x,y,t)$ means (input image, target class label, input task identity). The right side illustrates the neural network model and the predicted $P(Y)$ of the model. The color of each bar in the categorical distribution maps to a specific output node in the classifier. Note that Split MNIST generates five splits in sequence (0/1, 2/3, 4/5, 6/7, 8/9) for task $T_1$ to $T_5$ while here we only demonstrate the differences between $T_1$ and $T_2$. }\label{fig:split_mnsit}
\end{figure}

\section{Generating Task Sequences for Evaluating Continual Learning}
\subsection{Existing Experimental Methodologies}

In order to evaluate continual learning methods, scenarios are commonly generated from datasets using two operations: permutation and splitting. The typical source dataset is MNIST \cite{lecun1998mnist}, an image dataset of hand-written digits. The Permuted MNIST experiment \cite{goodfellow2013empirical} involves ten-digit classification, where each task consists of different permutations of the pixels in the images. The number of different permutations represents the length of the task sequence. This evaluation scenario is widely adopted \cite{kirkpatrick2017overcoming, lee2017IMM, lopez2017GEM, nguyen2017variational, ritter2018online, shin2017DGR, zenke2017SI}, despite criticism that it is less challenging in terms of forgetting \cite{farquhar2018VGER}. Another typical scenario, the Split MNIST experiment, was initially introduced in a multi-headed form where the ten digits are split into five two-class classification tasks (the model has five output heads, one for each task) \cite{zenke2017SI, shin2017DGR, nguyen2017variational}, and the task identity (1 to 5) is given at testing time. This scenario is argued to be easier since the selection of output head is given by the task identity \cite{farquhar2018VGER}. Farquhar and Gal \cite{farquhar2018VGER} propose a single-headed variant which does not require task identity, where it always requires the model to make a prediction over all classes (digits 0 to 9). Such single-headed Split MNIST is known as incremental class learning \cite{rebuffi2017icarl, parisi2018continual, maltoni2018continuous}. Van and Tolias \cite{van2018generative} propose another variant of single-headed Split MNIST, where the model always predicts over two classes instead of ten classes. Furthermore, the similar multi-headed/single-headed strategies in Split MNIST can apply to Permuted MNIST \cite{van2018generative} resulting in many combinations. These  scenarios are used in different works, and therefore there is a lack of coherent comparison. This paper addresses this problem by providing a systematic interpretation of the differences between an old task and a new one (see Section \ref{sec:categorization}).

\subsection{A Categorization of Current Scenarios}\label{sec:categorization}

We now provide a uniform categorization for these different combinations. The challenge of continual learning comes from the differences between tasks. The differences can be described by a shifting input/output distribution, and whether the input/output share the same representation space. For notation, we use $T_1 = (X_1, Y_1, t_1)$ to represent the old task, where $X = \{x_i\}_{1 \leq i \leq n}$ is a set of learning samples, $Y=\{y_i\}_{1 \leq i \leq n}$ is a set of output targets, and $t \in \mathbb{Z}$ is a task identity shared among all samples in a task. We use $T_2 = (X_2, Y_2, t_2)$ to denote a new task. A task sequence, $\{T_1,T_2,...,T_k\}$, represents a scenario for continual learning. The types of differences between tasks are summarized in Table \ref{tbl:task_dif}. We categorize differences into the following (note that \cite{van2018generative} independently developed a similar categorization which we align with but more formally describe):

\zk{Shorten}
\textbf{Incremental Domain Learning:} First, we discuss change in the marginal probability distribution of inputs $P(X)$, specifically $P(X_1) \neq P(X_2)$. The input distribution (domain) difference has been discussed extensively in the setting of transfer learning \cite{pan2009survey}, primarily as a domain adaptation problem \cite{pan2011domain}. Unlike domain adaptation, which aims to transfer knowledge from an old task to a new task where only the performance of the new task is considered, the continual learning setting aims to keep performance on old tasks while achieving reasonable performance on the new one as well using a single model. 
\zk{Removed, is this necessary?: To create a difference $P(X_1) \neq P(X_2)$, a straightforward example with RGB-image is having nonzero values only in the red channel for images in $T_1$, while the same set of images in $T_2$ have nonzero values only in the green channel.}\yhnote{ok} This domain difference can be caused by the permutation and splitting strategies mentioned previously. In fact, the most widely adopted Permuted MNIST experiment \cite{kirkpatrick2017overcoming, lee2017IMM, lopez2017GEM, nguyen2017variational, ritter2018online, shin2017DGR, zenke2017SI} generates such a domain difference. However, the inputs generated by the random permutation protocol is highly uncorrelated \cite{zenke2017SI,farquhar2018VGER}; thus they can not represent all possible scenarios. To generate a scenario with better-correlated tasks, one should avoid random permutation to keep the original spatial correlation between image pixels, allowing the possibility of shared features among tasks. This requirement can be fulfilled by a variant of the Split MNIST experiment, where ten digits are split into five binary classification tasks and the model has only a binary classifier (single-headed). Such a scenario is illustrated in the middle column of Figure \ref{fig:split_mnsit}. The requirement of using a single-headed model is essential to control for other types of differences. Specifically, the single-headed model ensures the same output space $\{Y_1\}=\{Y_2\}$ (task identity becomes unnecessary), and the equal amount of MNIST digits ensures the output distributions of the binary classification are the same ($P(Y_1)=P(Y_2)$). As a result, the only difference in the middle column of Figure \ref{fig:split_mnsit} is that the input images for label $\{0,1\}$ shift from digit 0/1 to digit 2/3.

\textbf{Incremental Class Learning:} The second scenario relates to multiclass classification. Each task in the sequence contains an exclusive subset of classes in a dataset. $P(X_1) \neq P(X_2)$ is true by the nature of this setting. All class labels are in the same naming space (single-headed) and the number of output nodes equals the number of total classes in the task sequence. Due to the multiclass property, $P(Y_1) \neq P(Y_2)$ is a natural consequence\yhnote{Is it clear enough?}. The right-most column in Figure \ref{fig:split_mnsit} demonstrates how split-dataset strategy \cite{rebuffi2017icarl,farquhar2018VGER} generates this scenario. The permutation strategy can also generate the task sequence \cite{van2018generative}. In the latter case, each permuted digit represents a new class, so the total number of classes is multiplied by the number of permutations, e.g., total $10 \times 10$ classes in a ten Permuted MNIST experiment.

\textbf{Incremental Task Learning:}
In the last scenario, the output spaces are disjoint between tasks, denoted as $\{Y_1\} \neq \{Y_2\}$. This definition makes $P(Y_1) \neq P(Y_2)$ true as well, while $P(X_1) \neq P(X_2)$ is generally true since the semantic classes differ. The differences between the output spaces are the output dimension and their associated semantic meaning. For example, the old task can be a classification problem of 5 classes while the new task is a regression task of a single value\zk{This is never really the case in the experiments (classification to regression)?}. To allow a model to produce an output for a specific task, a model requires task-specific output components which are selected by additional information, the task identifier $t$. A typical neural network for this scenario has a multi-headed output layer (one head for each task) \cite{zenke2017SI}. At testing time, only the head matching the $t$ will be activated to make predictions. One common approach to generate this scenario is having multiple datasets (ex: MNIST, CIFAR10, SVHN, AudioSet, CUB-200, etc.) and use one of them in one task \cite{aljundi_MAS, HAT, AAAI_measure, kemker2017fearnet}. The splitting and permutation strategies can also generate task sequences for this scenario, illustrated in Figure \ref{fig:split_mnsit} and Appendix Figure \ref{fig:permuted_mnsit}. The prior works mentioned in this paragraph commonly have the task identity given during testing; thus the experiments in this work follow the same setting.

\section{Experiments}
\label{sec:exp}
\begin{table}[]
\centering
\caption{The average accuracy (\%, higher is better) of all seen tasks after learning the task sequence generated by Split MNIST (Figure \ref{fig:split_mnsit}). The \textit{Memory} column means whether a method uses a memory mechanism, which further divides the methods into two groups in the comparison. The total static memory overhead is controlled to be the same among L2, Naive rehearsal, Naive rehearsal-C, online EWC, SI, MAS, GEM, and DGR. Each value is the average of 10 runs.}
\label{tbl:split_mnist_exp}
\begin{tabular}{cccccc}
\toprule
& \multirow{2}{*}{Method} & \multirow{2}{*}{Memory} & Incremental   & Incremental     & Incremental    \\
&                         & & task learning & domain learning & class learning \\ \toprule
\multirow{6}{*}{Baselines}& Adam &                    & 93.46 $\pm$ 2.01          & 55.16 $\pm$ 1.38            & 19.71 $\pm$ 0.08           \\
& SGD                     & & 97.98 $\pm$ 0.09          & 63.20 $\pm$ 0.35            & 19.46 $\pm$ 0.04           \\
& Adagrad                 & & 98.06 $\pm$ 0.53          & 58.08 $\pm$ 1.06            & 19.82 $\pm$ 0.09           \\ 
& L2                      & & 98.18 $\pm$ 0.96          & 66.00 $\pm$ 3.73            & 22.52 $\pm$ 1.08           \\ \arrayrulecolor{gray}\cmidrule{2-6}\arrayrulecolor{black}
& Naive rehearsal         & \checkmark & 99.40 $\pm$ 0.08          & 95.16 $\pm$ 0.49            & 90.78 $\pm$ 0.85           \\
& Naive rehearsal-C       & \checkmark & \textbf{99.57} $\pm$ 0.07          & \textbf{97.11} $\pm$ 0.34            & \textbf{95.59} $\pm$ 0.49           \\ \midrule
\multirow{8}{1.4cm}{\centering Continual learning methods} & EWC                & & 97.70 $\pm$ 0.81          & 58.85 $\pm$ 2.59            & 19.80 $\pm$ 0.05               \\ 
& Online EWC              & & 98.04 $\pm$ 1.10          & 57.33 $\pm$ 1.44            & 19.77 $\pm$ 0.04               \\ 
& SI                      & & 98.56 $\pm$ 0.49          & 64.76 $\pm$ 3.09            & 19.67 $\pm$ 0.09           \\
& MAS                     & & 99.22 $\pm$ 0.21          & 68.57 $\pm$ 6.85            & 19.52 $\pm$ 0.29           \\
& LwF                     & & 99.60 $\pm$ 0.03          & 71.02 $\pm$ 1.26            & 24.17 $\pm$ 0.33           \\ \arrayrulecolor{gray}\cmidrule{2-6}\arrayrulecolor{black}
& GEM                    & \checkmark & 98.42 $\pm$ 0.10          & 96.16 $\pm$ 0.35            & 92.20 $\pm$ 0.12           \\
& DGR                    & \checkmark & 99.47 $\pm$ 0.03          & 95.74 $\pm$ 0.23            & 91.24 $\pm$ 0.33           \\
& RtF                     & \checkmark & \textbf{99.66} $\pm$ 0.03          & \textbf{97.31} $\pm$ 0.11            & \textbf{92.56} $\pm$ 0.21           \\ \midrule
\multicolumn{2}{c}{Offline (upper bound)}   & & 99.52 $\pm$ 0.16          & 98.59 $\pm$ 0.15            & 97.53 $\pm$ 0.30          \\
\bottomrule
\end{tabular}
\end{table}

We now describe the experimental configuration. Here, we use the MNIST dataset with the splitting strategy (Figure \ref{fig:split_mnsit}) to generate the three continual learning scenarios in Table \ref{tbl:task_dif}. The standard train/test split was used, with 60k training images ($\sim$6k per digit) and 10k test images ($\sim$1k per digit). The preprocessing of images includes zero padding to 32x32 pixels and a standard normalization to zero mean with unit variance. No other data augmentation, (e.g. random translation) is applied. 

For a fair comparison, all methods use the same neural network architecture, which is a multi-layered perceptron with two hidden layers of 400 nodes each, followed by a softmax output layer. Both hidden layers use ReLU for the activation function. The loss function is a standard cross-entropy for classification in all methods and scenarios. All models are trained for 4 epochs per task with mini-batch size 128 using the Adam optimizer ($\beta_1=0.9$, $\beta_2=0.999$, learning rate$=0.001$) as the default unless explicitly described. In all experiments, the optimizer is never reset.

We use three baseline strategies. The most common baseline used in prior work is a neural network sequentially trained on all tasks in the standard way, in that the \textbf{parameters learned from old tasks are fine-tuned to the new task}. Such a model is usually optimized with Adam \cite{kingma2014adam}, but here we use different optimizers including Adam, SGD, and Adagrad \cite{Adagrad}. In fact, we show that Adam is in general a poor choice for this task. The latter two optimizers use 0.01 for the learning rate without momentum in all scenarios. Another baseline, \textbf{L2 regularization}, prevents the parameters from deviating too much from those previously learned. Note that this is similar to EWC in that the identity matrix replaces the Fisher information matrix. In other work this is only evaluated in one specific scenario (permuted MNIST) with limited length (3) of task sequence \cite{kirkpatrick2017overcoming}. Similar to other regularization methods, the L2 regularization requires tuning of the single-valued regularization coefficient, which is done by a grid search \cite{kirkpatrick2017overcoming, zenke2017SI}\zk{(see Appendix Section XXX for EWC and SI sensitivity)}\yhnote{added ref}. 


The third baseline is a \textbf{naive rehearsal strategy}, which is sometimes called experience replay. The model has a small replay buffer to store a fraction of previous data randomly. While training a new task, each mini-batch is constructed by an equal amount (64/64) of new data and the rehearsal data. The buffer size is predefined and fixed to match the space overhead introduced by online EWC and SI (\#parameter $\approx (1024 \times 400+400 \times 400)\times 2 = 1,139,200$, ignoring small overhead), which converts to 1.1k images when the pixel value is saved in a 32-bit floating number (named Naive rehearsal). Additionally, with the same memory space, it can store more images when compression is used. One naive compression is using an 8-bit integer to represent the pixel value; thus 4.4k images can be stored (named Naive rehearsal-C). For the buffer management, all tasks seen so far have an equal amount of images in the buffer while keeping the total number the same. This management is similar to iCaRL \cite{rebuffi2017icarl}, except that we randomly pick the images for staying in the buffer.

For comparison, we pick several popular methods (EWC\cite{kirkpatrick2017overcoming}, online EWC\cite{Schwarz18}, SI\cite{zenke2017SI}, MAS\cite{MAS}, LwF\cite{lwf}, GEM\cite{lopez2017GEM}) and state-of-the-art rehearsal-based methods (DGR\cite{shin2017DGR}, RtF\cite{van2018generative}) with generative model. The hyperparameter is tuned by a grid search, and the results with the best setting are reported. The total static memory overhead is controlled to be the same among Naive rehearsal, Naive rehearsal-C, L2, online EWC, SI, MAS, GEM, and DGR. RtF has only half the overhead of DGR since its classification model is shared with its generative model. The hyperparameter of EWC, online EWC, SI, and MAS is tuned with grid search. We use the results from Ven and Tolias \cite{van2018generative}, which provides an analysis and comparison developed concurrently with our work, for LwF, DGR, and RtF since the same model and training procedures are used.

\section{Results and Discussion}

Three interesting points can be seen in the results in Table \ref{tbl:split_mnist_exp}. First, Adagrad and L2 achieve better performance than online EWC and similar to SI. This shows that while Adam is popularly used for this task, Adagrad is more appropriate. This is possibly due to the fact that it results in small updates for parameters frequently used for past tasks. Second, naive rehearsal achieves performance similar to state-of-the-art methods with the same space overhead, and performs much better than online EWC and SI, especially in the incremental class scenario. This highlights the limitation of regularization-based methods and raises a question about the benefit of using a generative model, which is more difficult to train. The third is the obvious trend of difficulty among the three scenarios. The easiest one is incremental task learning, while the incremental class learning is harder than the incremental domain learning\yhnote{removed:, in that the incremental class scenario reaches a reasonable performance only with rehearsal-based methods}. 

A similar trend also happens in the scenarios generated by the permutation strategy, which can be seen in Appendix Table \ref{tbl:permuted_mnist_exp}. In that case, SI and online EWC are significantly better than Adagrad in only one of the three scenarios (incremental class), although all three methods present poor performance in that scenario. One aspect that is not apparent from these results, but is illustrated in Appendix Figure~\ref{fig:sen_reg}, is that EWC and SI variants require significant hyper-parameter tuning, with a wide gap between their worst and best performance (no such tuning was done for Adagrad). This would be difficult to tune automatically in real-world scenarios, however. One interesting cross-table comparison is that the performance of our six baselines in the scenarios generated by permutation (Appendix Table \ref{tbl:permuted_mnist_exp}) is generally comparable or higher than the same scenarios generated by splitting (Table \ref{tbl:split_mnist_exp}), although the Permuted MNIST scenarios have a larger number of classes and tasks. Such a result indicates that the permutation strategy creates simpler scenarios.

The last highlight is the comparison between Adagrad and EWC. We note that the difference between their performance is not significant in most of the experimental settings, yet EWC requires knowing the boundaries of a task to calculate Fisher information and store the parameters before switching to the next task. This creates a requirement that makes EWC less applicable than Adagrad in a real-world scenario, where the task boundaries are usually not available. Other regularization-based methods, such as SI and MAS, also suffer from the same limitation.

The strong baseline performance, especially in the incremental task learning\zk{actually all scenarios?}\yhnote{better?}, does not mean a scenario is solved. One can easily increase the difficulty by using a harder dataset or using a set of datasets to extend the length of a task sequence, as demonstrated in Appendix Section \ref{long_queue}. Under longer task sequences, regularization-based methods continue to degrade over time leading to questions about whether they fundamentally address catastrophic forgetting. Indeed, biological systems continue to learn new tasks with very little degradation in performance, even when that task has not been seen for a while. One avenue of research in this respect is to take a closer look at continual learning for tasks where feature sharing is possible. How such shared features can be learned when possible, or augmented with new features when tasks distributions differ significantly, is an open question. We also argue that more future effort should be put into scenarios that do not require knowing the task identity (incremental domain/class). Such scenarios are not only harder but also closer to a real scenario where the prior information about task selection is usually weak.

\subsubsection*{Acknowledgments}
This research is supported by DARPA’s Lifelong Learning Machines (L2M) program, under Cooperative Agreement HR0011-18-2-001.

\small
\bibliographystyle{unsrt}
\bibliography{ref}
\normalsize

\newpage
\appendix
\section*{Appendices}
\addcontentsline{toc}{section}{Appendices}
\input{appendix}

\end{document}

%% file: appendix.tex
\section{Permuted MNIST Experiments} \label{sec:permuted_mnist}

\begin{figure}[h]
  \centerfloat
  \includegraphics[clip, trim=0cm 1.3cm 0cm 1.7cm, width=1.\textwidth]{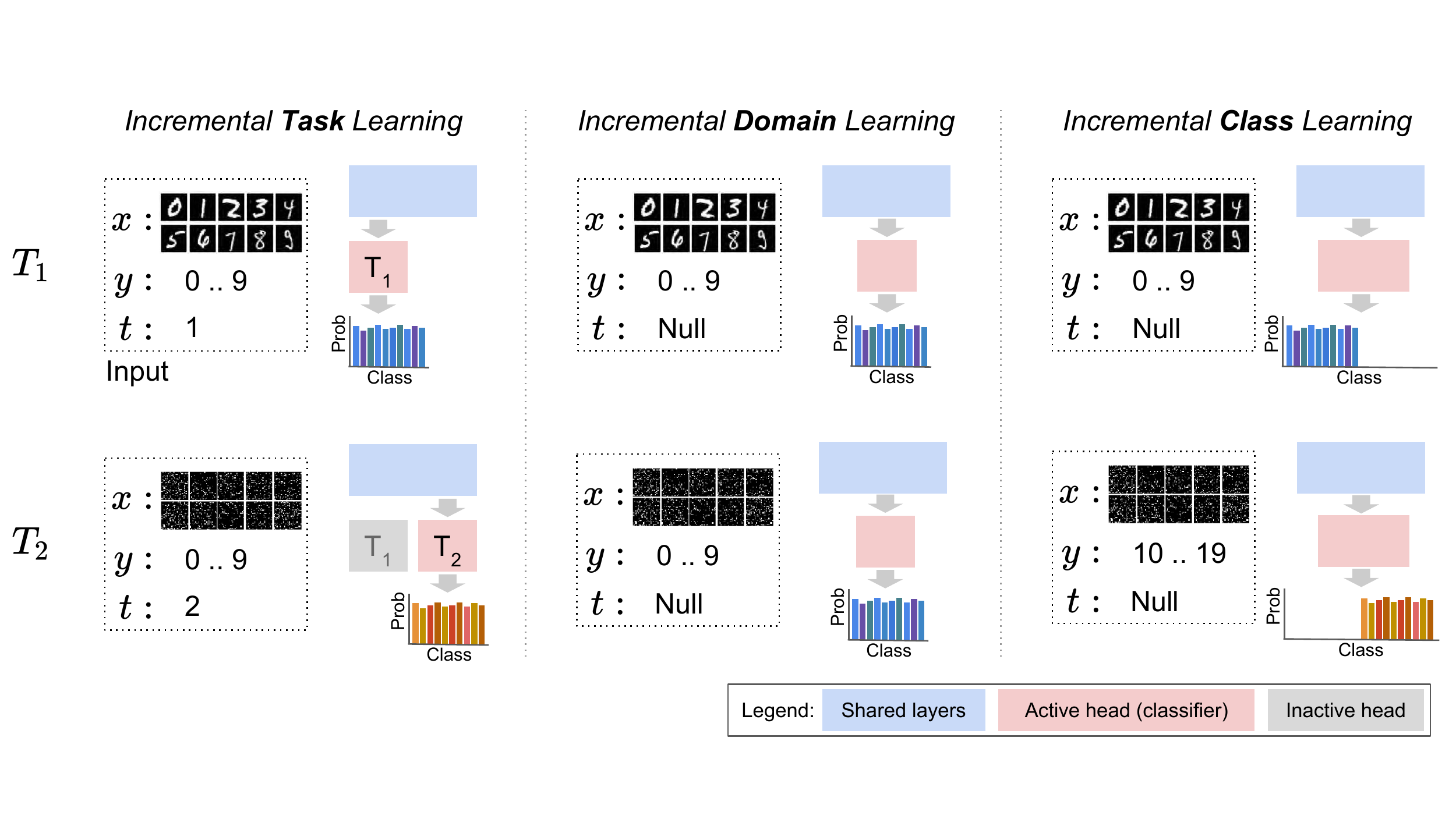} 
  \caption{The three continual learning scenarios generated by Permuted MNIST. In each sub-figure, the left dotted rectangle represents the inputs for training, in that $(x,y,t)$ means (input image, target class label, input task identity). The right side illustrates the neural network model and the predicted $P(Y)$ of the model. The color of each bar in the categorical distribution maps to a specific output node in the classifier. Note that the task sequence has 10 different permutations for task $T_1$ to $T_{10}$ while here we only demonstrate the differences between $T_1$ and $T_2$. }\label{fig:permuted_mnsit}
\end{figure}

In the Permuted MNIST experiments, the pre-processing of an image is similar to Split MNIST except that the order of pixels are permuted. The neural network architecture is also similar to the one in Split MNIST, except that the number of nodes in both hidden layers is 1000. Since the network size is larger, the space overhead introduced in online EWC and SI is larger (\#parameter $\approx (1024 \times 1000+1000 \times 1000) \times 2 = 4,048,000$) and converts to a buffer of 4k images in Naive rehearsal, or 16k compressed images in Naive rehearsal-C. The generative model in DGR (implemented by \cite{van2018generative}) uses a variational autoencoder whose encoder has the same architecture as the classification model; therefore DGR has a similar space overhead as online EWC and SI. In all scenarios, the standard cross-entropy for classification is optimized for 10 epochs with a learning rate ten-times smaller than Split MNIST experiments. Here we use 0.0001 for Adam, and 0.001 for SGD and Adagrad.

The best regularization coefficient for L2, EWC, online EWC, SI, and MAS is obtained through a grid search in each scenario. Our results in Table \ref{tbl:permuted_mnist_exp} is averaged from ten runs with random neural network initialization (include randomly initialized parameters in network heads which leads to much better baseline performance, compared to the zero initialization used in \cite{zenke2017SI}). Note that the results of LwF, DGR, and RtF are from \cite{van2018generative}, which uses the same neural network architecture and training procedure. 

Figure \ref{fig:permuted_mnsit} illustrated how to use the permutation strategy to generate the three learning scenarios. Note that the number of classes here is larger than the scenarios generated by the splitting strategy (incremental task/domain: 10 versus 2; incremental class: 100 versus 10).

Table \ref{tbl:incremental_class_two_case} compares two different initialization strategies in the hardest scenario, the incremental class learning. The two initialization strategies have the output nodes of all classes pre-allocated or not. In the pre-allocated setting, which has all output nodes be subject to the classification loss right from the beginning, an output node firstly sees negative samples since the first task, followed by some positive samples (in its corresponding task that contains the class), then again sees negative samples in the remaining tasks till the end. In the setting without pre-allocating the output nodes (the default used in our Table \ref{tbl:split_mnist_exp} and \ref{tbl:permuted_mnist_exp} as well as prior works), an output node is created while a new class arrived; thus the output node firstly sees some positive samples (in the newly arrived task), followed by negative samples in the remaining tasks. The results show that the scenario becomes easier with pre-allocating which enables the output nodes to learn from the beginning of the scenario.


\begin{table}[H]
\centering
\caption{The average accuracy (\%, higher is better) of all seen tasks after learning with the task sequence generated by Permuted MNIST (Figure \ref{fig:permuted_mnsit}). The \textit{Memory} column means whether a method uses a memory mechanism, which further divides the methods into two groups in the comparison. The total static memory overhead is controlled to be the same among L2, Naive rehearsal, Naive rehearsal-C, online EWC, SI, MAS, GEM, and DGR. Each value is the average of 10 runs.}
\label{tbl:permuted_mnist_exp}
\begin{tabular}{cccccc}
\toprule
& \multirow{2}{*}{Method} & \multirow{2}{*}{Memory} & Incremental   & Incremental     & Incremental    \\
&                         & & task learning & domain learning & class learning \\ \toprule
\multirow{6}{*}{Baselines} & Adam &                    & 93.42 $\pm$ 0.56              & 74.12 $\pm$ 0.86                & 14.02 $\pm$ 1.25              \\
& SGD                     & & 94.74 $\pm$ 0.24              & 84.56 $\pm$ 0.82                & 12.82 $\pm$ 0.95               \\
& Adagrad                 & & 94.78 $\pm$ 0.18              & 91.98 $\pm$ 0.63                & 29.09 $\pm$ 1.48               \\
& L2                      & & 95.45 $\pm$ 0.44              & 91.08 $\pm$ 0.72                & 13.92 $\pm$ 1.79               \\ \arrayrulecolor{gray}\cmidrule{2-6}\arrayrulecolor{black}
& Naive rehearsal         & \checkmark & 96.67 $\pm$ 0.12                 & 95.19 $\pm$ 0.11                & 96.25 $\pm$ 0.10               \\
& Naive rehearsal-C       & \checkmark & \textbf{97.36} $\pm$ 0.03                            & \textbf{96.28} $\pm$ 0.47                & \textbf{97.24} $\pm$ 0.05               \\ \midrule
\multirow{8}{1.4cm}{\centering Continual learning methods} & EWC                & & 95.38 $\pm$ 0.33               & 91.04 $\pm$ 0.48                  & 26.32 $\pm$ 4.32            \\ 
& Online EWC              & & 95.15 $\pm$ 0.49              & 92.51 $\pm$ 0.39                & 42.58 $\pm$ 6.50            \\ 
& SI                      & & 96.31 $\pm$ 0.19              & 93.94 $\pm$ 0.45                & 58.52 $\pm$ 4.20               \\
& MAS                     & & 96.65 $\pm$ 0.18              & 94.08 $\pm$ 0.43                & 50.81 $\pm$ 2.92               \\
& LwF                     & & 69.84 $\pm$ 0.46              & 72.64 $\pm$ 0.52                & 22.64 $\pm$ 0.23               \\ \arrayrulecolor{gray}\cmidrule{2-6}\arrayrulecolor{black}
& GEM                     & \checkmark & 97.05 $\pm$ 0.07              & 96.19 $\pm$ 0.11                & \textbf{96.72} $\pm$ 0.03               \\
& DGR                     & \checkmark & 92.52 $\pm$ 0.08              & 95.09 $\pm$ 0.04                & 92.19 $\pm$ 0.09               \\
& RtF                     & \checkmark & \textbf{97.31} $\pm$ 0.01              & \textbf{97.06} $\pm$ 0.02                & 96.23 $\pm$ 0.04               \\ \midrule
\multicolumn{2}{c}{Offline (upper bound)}   &  & 98.01 $\pm$ 0.04              & 97.90 $\pm$ 0.09                & 97.95 $\pm$ 0.04               \\
\bottomrule
\end{tabular}
\end{table}

\begin{table}[H]
\centering
\caption{The average accuracy (\%, higher is better) of two \textbf{Incremental Class Learning variants}. The first case has an unknown number of total classes (the columns with \textit{No}). The number of output nodes increases along with the total number of seen classes. During training, only the output nodes of seen classes are subject to the classification loss. This is the setting used in Table \ref{tbl:split_mnist_exp} and \ref{tbl:permuted_mnist_exp}. The second case has a known number of total classes (the columns with \textit{Yes}). All of the output nodes are pre-allocated and are subject to the classification loss since the first task. The results indicate that the incremental class scenario generated by Permuted MNIST is much easier when the total number of classes is known. In contrast, the scenario generated by Split MNIST has a similar difficulty between the two variants.}
\label{tbl:incremental_class_two_case}
\begin{tabular}{cc|cc|cc}
\toprule
                            &                    & \multicolumn{2}{c|}{Split MNIST} & \multicolumn{2}{c}{Permuted MNIST} \\
\multicolumn{2}{c|}{Known Total \#class?}         & \textit{No}   & \textit{Yes}  & \textit{No}    & \textit{Yes}    \\
\toprule
\multirow{6}{*}{Baselines}   & Adam               & 19.71 $\pm$ 0.08   & 19.67 $\pm$ 0.05   & 14.02 $\pm$ 1.25    & 42.32 $\pm$ 3.22    \\
                            & SGD                 & 19.46 $\pm$ 0.04   & 19.44 $\pm$ 0.03   & 12.82 $\pm$ 0.95    & 17.54 $\pm$ 0.81    \\
                            & Adagrad             & 19.82 $\pm$ 0.09   & 19.75 $\pm$ 0.08   & 29.09 $\pm$ 1.48    & 79.50 $\pm$ 3.70    \\
                            & L2                  & 22.52 $\pm$ 1.08   & 20.54 $\pm$ 1.12   & 13.92 $\pm$ 1.79    & 43.18 $\pm$ 2.30    \\
                            & Naive rehearsal     & 90.78 $\pm$ 0.85   & 89.64 $\pm$ 0.63   & 96.25 $\pm$ 0.10    & 96.24 $\pm$ 0.11    \\
                            & Naive rehearsal-C   & 95.59 $\pm$ 0.49   & 94.35 $\pm$ 0.20   & 97.24 $\pm$ 0.05    & 97.15 $\pm$ 0.05    \\ \midrule
\multirow{4}{1.4cm}{\centering Continual learning methods} & EWC         & 19.80 $\pm$ 0.05   & 19.76 $\pm$ 0.05   & 26.32 $\pm$ 4.32     & 91.95 $\pm$ 1.04    \\
                            & Online EWC          & 19.77 $\pm$ 0.04   & 19.71 $\pm$ 0.06   & 42.58 $\pm$ 6.50    & 86.57 $\pm$ 3.52    \\
                            & SI                  & 19.67 $\pm$ 0.09   & 20.88 $\pm$ 0.96   & 58.52 $\pm$ 4.20    & 79.36 $\pm$ 2.42    \\
                            & MAS                 & 19.52 $\pm$ 0.29   & 19.98 $\pm$ 0.34   & 50.81 $\pm$ 2.92    & 73.82 $\pm$ 1.67    \\
\bottomrule
\end{tabular}
\end{table}

\section{Lengthened Task Queue Experiments} \label{long_queue}
Most continual learning methods examine their models with less than $20$ tasks in the queue. A more challenging yet practical case of continual learning is to have a dynamic environment where varied levels of domain shifting are encountered; thus we augment the incremental task learning in Section~\ref{sec:exp} by extending the length of task queue from $5$ to $78$ with five datasets, including \textit{MNIST}, \textit{Fashion MNIST}, \textit{EMNIST letter}, \textit{SVHN}, and \textit{CIFAR100}. Each task only contains two classes, while each class presents only once in the task queue. 

The evaluation uses two different neural network architectures, CNN and MLP, to enrich the comparison, and we list the detailed architecture in Table ~\ref{tab:arch}. For a fair comparison, the number of parameters is similar ($\sim$300K parameters) between CNN and MLP. In the training stage, we adopt Adam as optimizer with a learning rate of 0.001 to train 10 epochs for each task with a batch size of 128. The learning rate for Adagrad is 0.001.

\begin{figure*}[t]
   \begin{picture}(0,250)
     \put(15,10){\includegraphics[width=7cm]{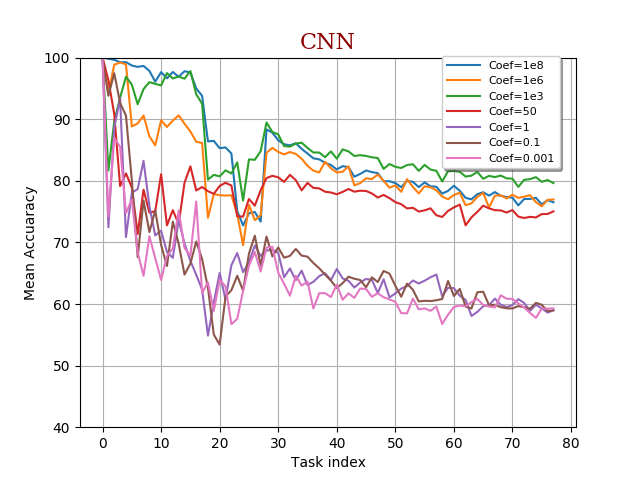}}
     \put(200,10){\includegraphics[width=7cm]{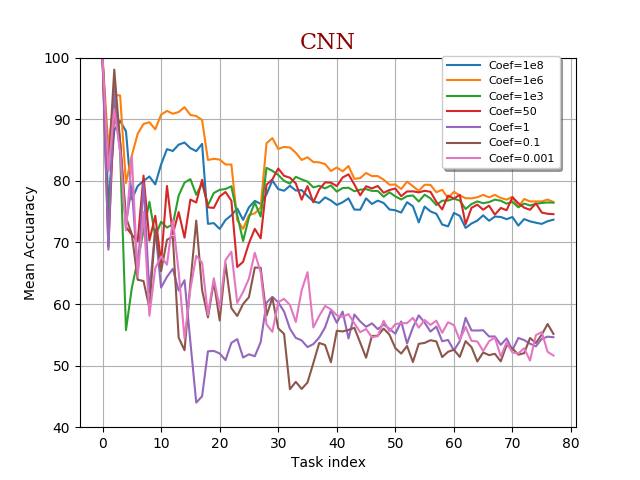}}
     \put(15,143){\includegraphics[width=7cm]{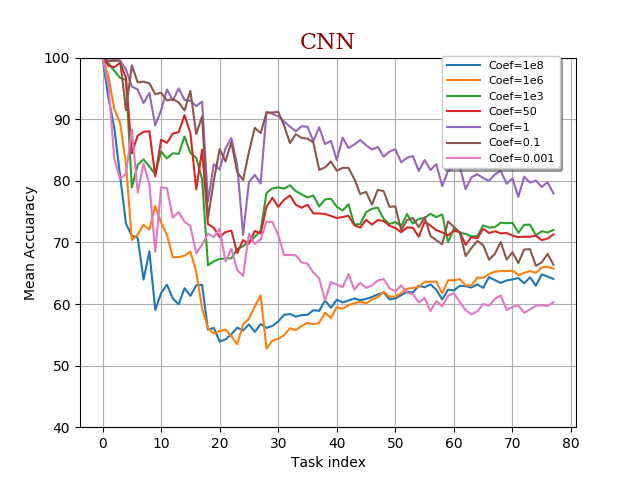}}
     \put(200,143){\includegraphics[width=7cm]{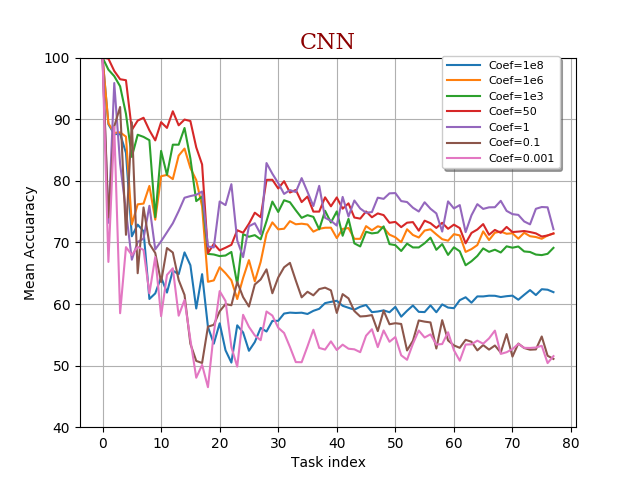}}
     \put(72,290){Random Initialization}
     \put(265,290){Zero Initialization}
     \put(5,63){ \rotatebox{90}{Online EWC}}
     \put(5,218){ \rotatebox{90}{SI}}
   \end{picture}
\vspace{-5mm}
\caption{ \textbf{Sensitivity to regularization weight.} Top row represents the results of SI, and the bottom row represents the results of Online EWC. Different initialization methods are used in different column.\yhnote{for online EWC, does a coef larger larger than 50 tried?}\yhnote{Is it only the heads are initialized differently? (That's what I do in the repo)}}
\label{fig:sen_reg}
\end{figure*}

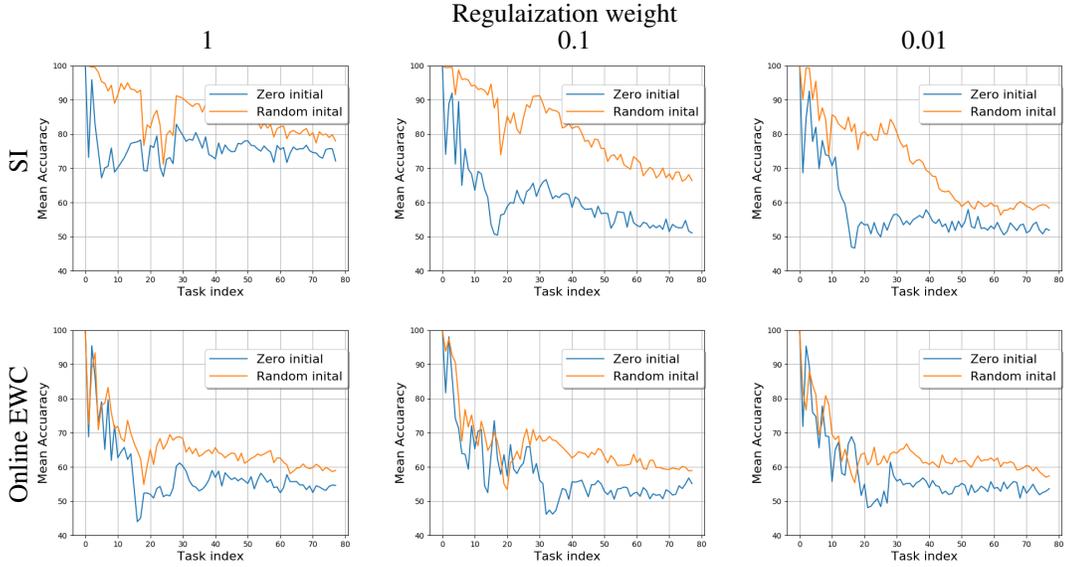
\begin{figure*}[]
   \begin{picture}(0,170)
     \put(15,10){\includegraphics[width=4.7cm]{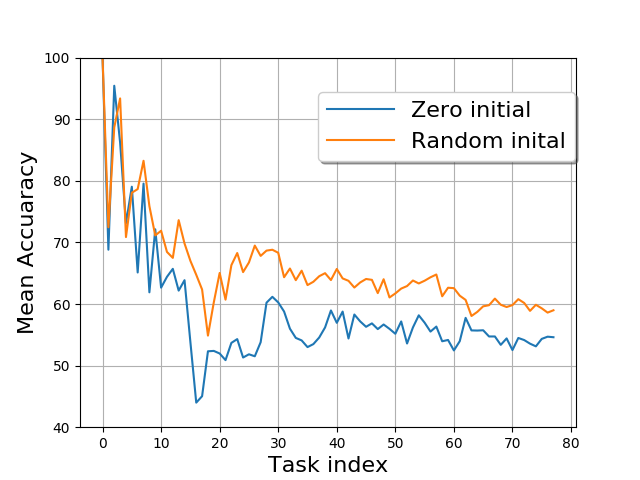}}
     \put(150,10){\includegraphics[width=4.7cm]{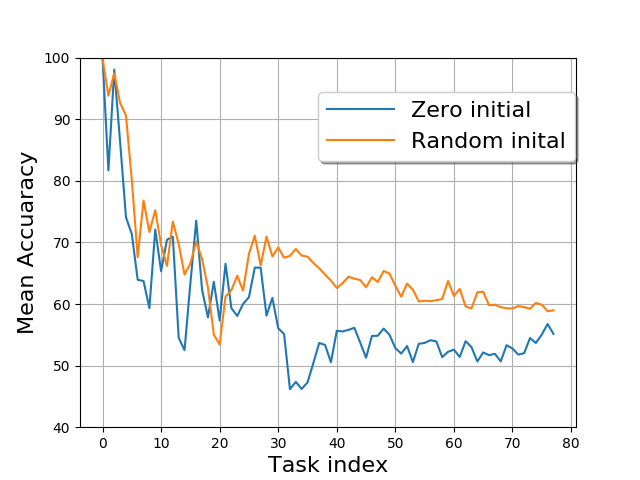}}
     \put(285,10){\includegraphics[width=4.7cm]{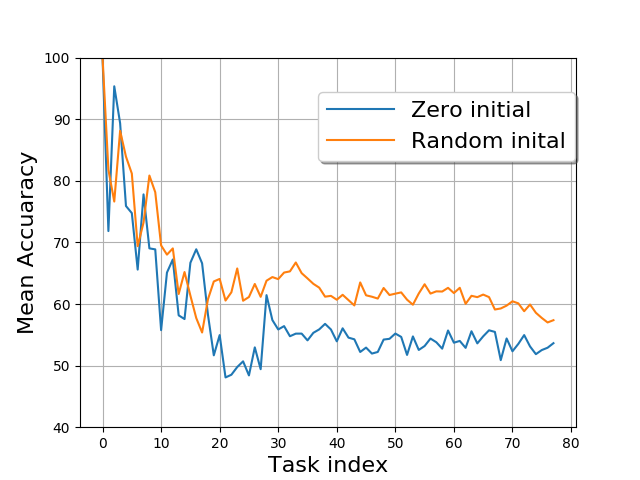}}
     \put(15,110){\includegraphics[width=4.7cm]{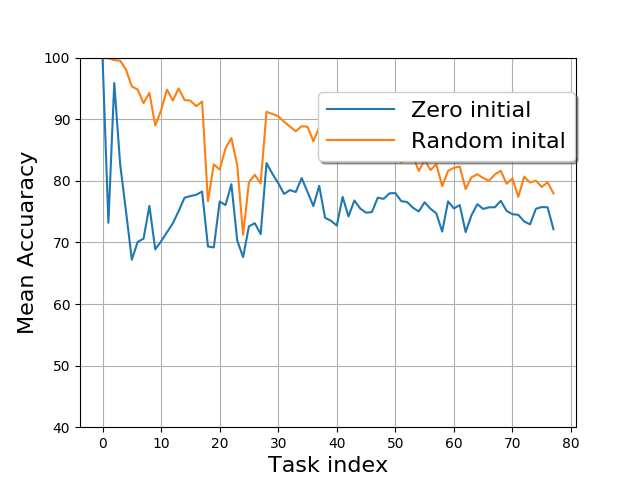}}
     \put(150,110){\includegraphics[width=4.7cm]{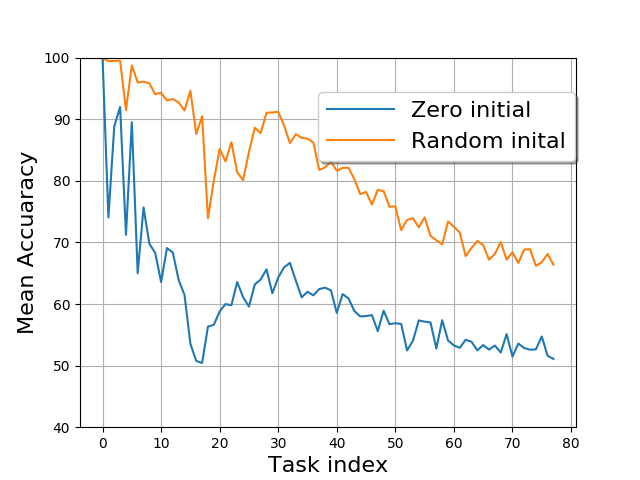}}
     \put(285,110){\includegraphics[width=4.7cm]{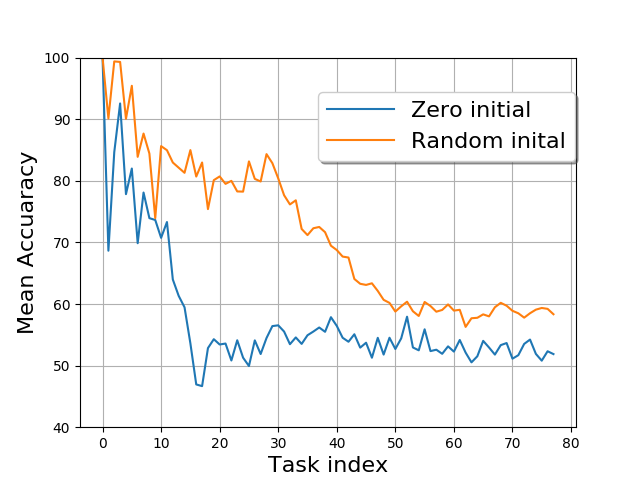}}
     \put(175,215){Regulaization weight}
     \put(80,205){1}
     \put(215,205){0.1}     
     \put(345,205){0.01}
     \put(5,33){ \rotatebox{90}{Online EWC}}
     \put(5,158){ \rotatebox{90}{SI}}
   \end{picture}
\vspace{-5mm}
\caption{ \textbf{Sensitivity to initialization method.} Top row represents the results of SI, and the bottom row represents the results of Online EWC. Different regularization weights are used in different columns.}
\label{fig:sen_init}
\end{figure*}

To examine the robustness of the regularization-based methods, we include SI and online EWC in the scenario of longer task queue. The results are presented in Table \ref{tbl:avg_mlp} (MLP) and \ref{tbl:avg_cnn} (CNN), demonstrating that the regularization-based methods can be worse than the baseline (Adam) if the regularization coefficient is not in tuned well. In contrast, the Adagrad achieves a similar level of performance without any hyperparameter search. The sensitivity analysis is provided in Figure \ref{fig:sen_reg} and \ref{fig:sen_init}, which shows that regularization-based methods are prone to the choice of the regularization coefficient and are sensitive to how the parameters (of the heads) been initialized.

\begin{figure}[]
   \begin{picture}(0,160)
     \put(0,10){\includegraphics[width=7.7cm]{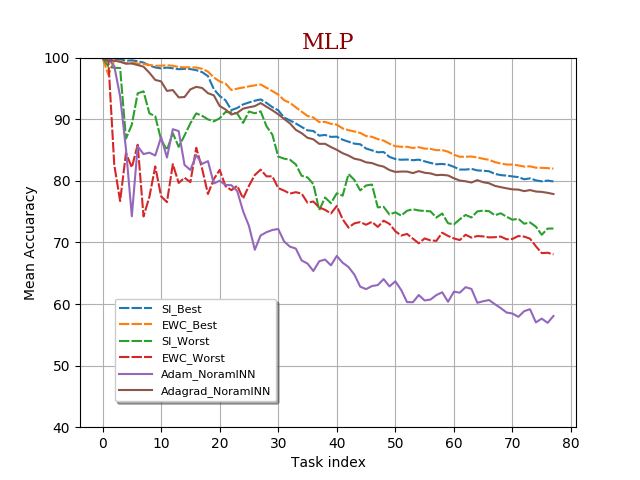}}
     \put(200,10){\includegraphics[width=7.7cm]{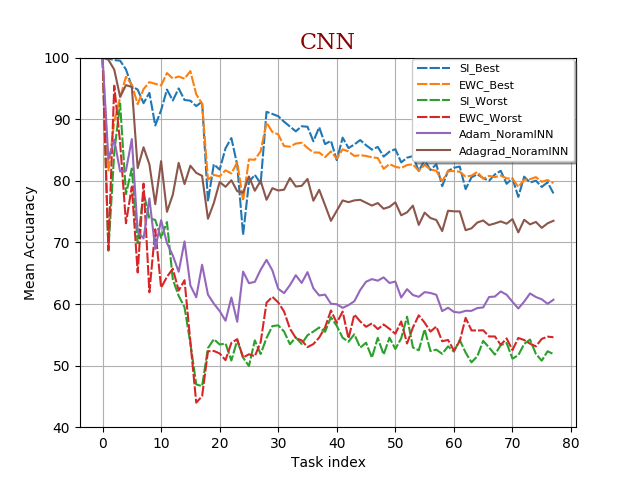}}
   \end{picture}
\vspace{-5mm}
\caption{ \textbf{A comparison between MLP and CNN models}. In each subfigure, we list SI and Online EWC with best and worst hyper-parameter selections (solid line) and two additional optimization methods (dashed line).}
\label{fig:avg_cnn_mlp}
\end{figure}

\input{table/multi_datasets_CNNMLP.tex}

\input{table/arch.tex}

%% file: table/multi_datasets_CNNMLP.tex
\begin{table}[]
\centering
\caption{\textbf{Average test accuracy (\%) for long task queue with MLP.} Note that SI and Online EWC also apply Adam as optimizer.}
\label{tbl:avg_mlp}
\begin{tabular}{ccccccc}
\toprule
     & \multicolumn{6}{c}{MLP}  \\\cmidrule(lr){2-7}
Task & Adam  & \begin{tabular}[c]{@{}c@{}}SI\\ (Best)\end{tabular} & \begin{tabular}[c]{@{}c@{}}SI\\ (Worst)\end{tabular} & \begin{tabular}[c]{@{}c@{}}Online EWC\\ (Best)\end{tabular} & \begin{tabular}[c]{@{}c@{}}Online EWC\\ (Worst)\end{tabular} & Adagrad \\\midrule 
1    & 99.95 & 99.95& 99.91& 99.95& 99.81& 99.95   \\
11   & 87.09 & 98.56& 94.77& 98.69& 92.29& 96.14   \\
21   & 80.01 & 96.62& 90.46& 96.78& 85.51& 92.15   \\
31   & 72.18 & 94.06& 88.29& 94.00 & 80.81& 90.80   \\
41   & 67.80 & 88.76& 77.46& 89.12& 72.21& 85.03   \\
51   & 63.68 & 84.39& 71.41& 85.61& 66.82& 81.44   \\
61   & 62.00 & 82.48& 69.67& 84.23& 65.76& 80.41   \\
71   & 58.45 & 81.26& 67.04& 82.64& 62.69& 78.61   \\
78   & 58.05 & 80.52& 66.33& 82.06& 61.06& 77.85  \\\bottomrule
\end{tabular}
\end{table}

\begin{table}[]
\centering
\caption{\textbf{Average test accuracy (\%) for long task queue with CNN.} Note that SI and Online EWC also apply Adam as optimizer.}
\label{tbl:avg_cnn}
\begin{tabular}{ccccccc}
\toprule
\multicolumn{1}{l}{} & \multicolumn{6}{c}{CNN}\\\cmidrule(lr){2-7}
Task& Adam   & \begin{tabular}[c]{@{}c@{}}SI\\ (Best)\end{tabular} & \begin{tabular}[c]{@{}c@{}}SI\\ (Worst)\end{tabular} & \begin{tabular}[c]{@{}c@{}}Online EWC\\ (Best)\end{tabular} & \begin{tabular}[c]{@{}c@{}}Online EWC\\ (Worst)\end{tabular} & Adagrad \\\midrule
1 & 100.0  & 100.0& 100.0& 100.0& 99.95& 100.0   \\
11& 73.61  & 91.46& 70.74& 95.48& 62.65& 83.19   \\
21& 58.87  & 81.81& 53.42& 80.72& 51.96& 79.80   \\
31& 62.47  & 90.48& 56.54& 87.54& 60.26& 78.42   \\
41& 59.96  & 83.35& 56.45& 83.59& 56.94& 75.14   \\
51& 63.65  & 85.14& 52.70& 82.27& 55.17& 76.51   \\
61& 58.72  & 82.12& 52.27& 81.58& 52.46& 75.04   \\
71& 60.36  & 80.40& 51.12& 80.34& 52.52& 73.78   \\
78& 60.70  & 77.95& 51.87& 79.63& 54.61& 73.51   \\\bottomrule
\end{tabular}
\end{table}

%% file: table/arch.tex
\begin{table*}[h]
\centering
\caption{The network architecture of CNN and MLP. Note that the input of MLP is a vector flattened from an image of size $28\times28\times1$. }
\label{tab:arch}
\begin{tabular}{ c c c c}
\toprule
\multicolumn{4}{c}{\textbf{CNN}}                                                   \\  \midrule
Layer                    & Activation Size            & Activ. Fun.   & Max Pooling\\ \midrule
 Input                   & $28\times28\times1$        &      -        & -          \\ \midrule
$10\times5\times5$ Conv. & $14\times14\times10$       &  ReLU         & \checkmark \\ 
$20\times5\times5$ Conv. &  $7\times7\times20$        &  ReLU         & \checkmark \\ 
$40\times5\times5$ Conv. &  $3\times3\times40$        &  ReLU         &  -         \\ 
$70\times5\times5$ Conv. &  $3\times3\times70$        &  ReLU         &  -         \\
Dense Layer              & $256$                      &  ReLU         &  -         \\ 
Dense Layer              & $2$                        &    -          &  -         \\ \midrule\midrule
                                                  \multicolumn{4}{c}{\textbf{MLP}}                           \\ \midrule
 Input                & $784$                    &                 \\ \midrule
Dense Layer       & $256$                    &  ReLU         &  -         \\ 
Dense Layer       & $256$                    &  ReLU         &  -         \\ 
Dense Layer       & $2$                      &  -            &  -         \\ \bottomrule
\end{tabular}
\end{table*}